\documentclass[10pt]{article} 
\usepackage[preprint]{tmlr}

\usepackage{amsmath,amsfonts,bm}

\def\eqref#1{equation~\ref{#1}}

\def\1{\bm{1}}

\DeclareMathAlphabet{\mathsfit}{\encodingdefault}{\sfdefault}{m}{sl}
\SetMathAlphabet{\mathsfit}{bold}{\encodingdefault}{\sfdefault}{bx}{n}

\newcommand{\R}{\mathbb{R}}

\usepackage{hyperref}
\usepackage{graphicx}
\usepackage{tablefootnote}
\usepackage{url}

\title{Using Multimodal Foundation Models and Clustering for Improved Style Ambiguity Loss}

\author{\name James Baker \email jlbaker361@gmail.com \\
      \addr Department of Computer Science\\
      University of Maryland, Baltimore County}

\newcommand{\exampleimgtable}[4]{
\begin{center}
    \begin{table}[h]
        \centering
        \begin{tabular}{|c|c|c|c|c|}
        \hline
           Prompt  &  CLIP Based & K Means Text & K Means Image & DCGAN \\
           \hline
           \vbox{\hbox{\strut a painting} \hbox{\strut of a man} \hbox{\strut } \hbox{\strut }} & \includegraphics[scale=0.3]{appendix-#4-#1/ddpo-clip-#1_0.jpg} &
           \includegraphics[scale=0.3]{appendix-#4-#1/ddpo-kmeans-text-#1_0.jpg} &
           \includegraphics[scale=0.3]{appendix-#4-#1/ddpo-kmeans-image-#1_0.jpg} &
           \includegraphics[scale=0.3]{appendix-#4-#1/ddpo-dcgan-#1_0.jpg} \\
           
           \vbox{\hbox{\strut a picture of} \hbox{\strut a woman} \hbox{\strut } \hbox{\strut }} & \includegraphics[scale=0.3]{appendix-#4-#1/ddpo-clip-#1_10.jpg} &
           \includegraphics[scale=0.3]{appendix-#4-#1/ddpo-kmeans-text-#1_10.jpg} &
           \includegraphics[scale=0.3]{appendix-#4-#1/ddpo-kmeans-image-#1_10.jpg} &
           \includegraphics[scale=0.3]{appendix-#4-#1/ddpo-dcgan-#1_10.jpg} \\
           
           \vbox{\hbox{\strut a painting of} \hbox{\strut an animal} \hbox{\strut } \hbox{\strut }} & \includegraphics[scale=0.3]{appendix-#4-#1/ddpo-clip-#1_5.jpg} &
           \includegraphics[scale=0.3]{appendix-#4-#1/ddpo-kmeans-text-#1_5.jpg} &
           \includegraphics[scale=0.3]{appendix-#4-#1/ddpo-kmeans-image-#1_5.jpg} &
           \includegraphics[scale=0.3]{appendix-#4-#1/ddpo-dcgan-#1_5.jpg} \\
           
           \vbox{\hbox{\strut a painting } \hbox{\strut of nature } \hbox{\strut }} & \includegraphics[scale=0.3]{appendix-#4-#1/ddpo-clip-#1_3.jpg} &
           \includegraphics[scale=0.3]{appendix-#4-#1/ddpo-kmeans-text-#1_3.jpg} &
           \includegraphics[scale=0.3]{appendix-#4-#1/ddpo-kmeans-image-#1_3.jpg} &
           \includegraphics[scale=0.3]{appendix-#4-#1/ddpo-dcgan-#1_3.jpg} \\
           
           \vbox{\hbox{\strut a drawing } \hbox{\strut of a man} \hbox{\strut }} & \includegraphics[scale=0.3]{appendix-#4-#1/ddpo-clip-#1_18.jpg} &
           \includegraphics[scale=0.3]{appendix-#4-#1/ddpo-kmeans-text-#1_18.jpg} &
           \includegraphics[scale=0.3]{appendix-#4-#1/ddpo-kmeans-image-#1_18.jpg} &
           \includegraphics[scale=0.3]{appendix-#4-#1/ddpo-dcgan-#1_18.jpg} \\
           
           \vbox{\hbox{\strut a drawing} \hbox{\strut of a woman} \hbox{\strut }} & \includegraphics[scale=0.3]{appendix-#4-#1/ddpo-clip-#1_19.jpg} &
           \includegraphics[scale=0.3]{appendix-#4-#1/ddpo-kmeans-text-#1_19.jpg} &
           \includegraphics[scale=0.3]{appendix-#4-#1/ddpo-kmeans-image-#1_19.jpg} &
           \includegraphics[scale=0.3]{appendix-#4-#1/ddpo-dcgan-#1_19.jpg} \\

           \vbox{\hbox{\strut (no prompt)} \hbox{\strut } \hbox{\strut }} & \includegraphics[scale=0.3]{appendix-#4-#1/ddpo-clip-#1_8.jpg} &
           \includegraphics[scale=0.3]{appendix-#4-#1/ddpo-kmeans-text-#1_8.jpg} &
           \includegraphics[scale=0.3]{appendix-#4-#1/ddpo-kmeans-image-#1_8.jpg} &
           \includegraphics[scale=0.3]{appendix-#4-#1/ddpo-dcgan-#1_8.jpg} \\
           
        \hline
        \end{tabular}
        \caption{#2}
        \label{tab:#3}
    \end{table}
\end{center}
}

\def\month{07}  
\def\year{2024} 
\def\openreview{\url{https://openreview.net/forum?id=XXXX}} 

\begin{document}

\maketitle

\begin{abstract}
Teaching text-to-image models to be creative involves using style ambiguity loss, which requires a pretrained classifier.  In this work, we explore a new form of the style ambiguity training objective, used to approximate creativity, that does not require training a classifier or even a labeled dataset. We then train a diffusion model to maximize style ambiguity to imbue the diffusion model with creativity and find our new methods improve upon the traditional method, based on automated metrics for human judgment, while still maintaining creativity and novelty.
\end{abstract}

\section{Introduction}
With every new invention comes a new wave of possibilities. Humans have been making pictures since before recorded history, so its only natural that there would be interest in computational image generation. Artificially generating photographs that are indistinguishable from real ones has become so easy and effective that there is even concern over "deepfakes" being used for propaganda or illicit purposes \citep{democracypawelec2022}. On the other hand, generating images that look like art is a slightly different problem. The exact mathematical properties of what constitutes "quality" art is not as easy to quantify as other tasks, like classification accuracy, prediction error or whether a question was answered correctly. While machines can very easily be trained to mimic a dataset, humans like to be surprised by novelty, without feeling like they are being exposed to total randomness. A breakthrough was the invention of the Creative Adversarial Network \citep{ElgammalLEM17}, which used a style ambiguity loss to train a network to generate images that could not be classified as belonging to a particular style. However, GANs have largely been superseded by diffusion models \citep{luo2022understanding}, due to their far better results. Additionally, the style ambiguity loss requires a pretrained classifier. Every set of styles or concepts requires training a classifier before even training a model to generate images. Furthermore, training a classifier requires that the dataset be labeled correctly, and manually labeling a dataset is often even more expensive and time-consuming than training a model. To circumvent these issues, we propose using a  classifier that does not require any additional training and can be easily applied to any dataset, labeled or unlabeled. Our contributions are as follows:
\begin{itemize}
  \item We applied creative style ambiguity loss to diffusion models, which are easier to train and produce higher-quality images than GANs.
  \item We developed versatile CLIP-based and K-Means-based creative style ambiguity losses that do not require training a separate GAN-based style classifier.
  \item Empirically, we find our new creative style ambiguity loss can be used to tune a diffusion model to generate samples that are higher quality than the generated samples of a diffusion model trained with the pre-existing GAN-based style ambiguity loss
\end{itemize}

\section{Related Work}
\subsection{Creativity}

Creativity has been hard to define and quantify. Creative work has been formulated as work having novelty, in that it differs from other similar objects, and also utility, in that it still performs a function \citep{cropley}. For example, a Corinthian column has elaborate, interesting, unexpected adornments (novelty) but still holds up a building (utility). A distinction can also be made between "P-creativity", where the work is novel to the creator, and "H-creativity" where the work is novel to everyone \citep{Boden}. Computational techniques to be creative include using genetic algorithms \citep{DiPaola_2008}, reconstructing artifacts from novel collections of attributes \citep{iqbal2016digital}, and most relevantly to this work, using Generative Adversarial Networks \citep{ElgammalLEM17} with a style ambiguity loss.
\subsection{Computational Art}
One of the first algorithmic approaches dates back to the 1970s with the now primitive AARON \citep{mccorduck1991aaron}, which was initially only capable of drawing black and white sketches. Generative Adversarial Networks \citep{goodfellow2014generative}, or GANs, were some of the first models to be able to create complex, photorealistic images and seemed to have potential to be able to make art. Despite many problems with GANs, such as mode collapse and unstable training \citep{saxena2023generative}, GANs and further improvements \citep{arjovsky2017wasserstein, karras2019stylebased, karras2018progressive} were state of the art until the introduction of diffusion \citet{Sohl-DicksteinW15}. Diffusion models such as IMAGEN \citep{saharia2022photorealistic} and DALLE-3 \citep{BetkerImprovingIG} have attained widespread commercial success (and controversy) due to their widespread adoption.
\subsection{Reinforcement Learning}
Reinforcement learning (RL) is a method of training a model by having it take actions that generate a reward signal and change the environment, thus changing the impact and availability of future actions \cite{qiang2011rl}. RL has been used for tasks as diverse as playing board games \citep{Silver2017MasteringTG}, protein design \citep{lutz2023top}, self-driving vehicles \citep{kiran2021deep} and quantitative finance \citep{sahu2023overview}. Policy-gradient RL \citep{sutton1999policygradient} optimizes a policy \(\pi\) that chooses which action to take at any given timestep, as opposed to value-based methods that may use a heuristic to determine the optimal choice. Examples of policy gradient methods include Soft Actor Critic \citep{haarnoja2018soft}, Deep Deterministic Policy Gradient \citep{lillicrap2019ddpg} and Trust Region Policy Optimization \citep{schulman2017trust}. 

\section{Method}
\subsection{Model}
\subsubsection{Creative Adversarial Network} \label{can}
A Generative Adversarial Network, or GAN \citep{goodfellow2014generative}, consists of two models, a generator and a discriminator. The generator generates samples from noise, and the discriminator detects if the samples are drawn from the real data or generated. During training, the generator is trained to trick the discriminator into classifying generated images as real, and the discriminator is trained to classify images correctly. Given a generator \(G: \mathbb{R}^{noise} \rightarrow \mathbb{R}^{h \times w \times 3}\), a discriminator \(D: \mathbb{R}^{h \times w \times 3} \rightarrow [0,1]\) real images \(x \in \mathbb{R}^{h \times w \times 3}\), and noise \(\mathcal{Z} \in \mathbb{R}^{noise}\), the objective is:
\[\underset{G}{min} \: \underset{D}{max} \: \mathbb{E}_{x}[log(D(x)]+\mathbb{E}_{\mathcal{Z}}[log(1-D(G(\mathcal{Z}))]\]
At inference time, the generator is used to generate realistic samples. 
\citet{ElgammalLEM17} introduced the Creative Adversarial Network, or CAN, which was a DCGAN \citep{radford2016unsupervised} where the discriminator was also trained to classify real samples, minimizing the style classification loss. Given \(N\) classes of image (such as ukiyo-e, baroque, impressionism, etc.), the classification modules of the Discriminator \(D_C: \mathbb{R}^{h \times w \times 3} \rightarrow \mathbb{R}^{N}\) that returns a probability distribution over the \(N_s\) style classes for an image and the real labels \(\ell \in \mathbb{R}^{N}\), the style classification loss was:
\[L_{SL}=\mathbb{E}_{x,\ell} [\textbf{CE}(D_C(x),\ell)]\]
Where \(\textbf{CE}\) is the cross entropy function.

The generator was also trained to generate samples that could not be easily classified as belonging to one class. This stylistic ambiguity is a proxy for creativity or novelty. Given a vector \(U \in \mathbb{R}^{N}\), where each entry \(u_1, u_2,,,u_N = \frac{1}{N}\), and some classifier \(C:\mathbb{R}^{h \times w \times 3} \rightarrow \mathbb{R}^{N}\)  the style ambiguity loss is:
\[L_{SA} = \mathbb{E}_{\mathcal{Z}} [\textbf{CE}(C(G(\mathcal{Z})),U)]\]
The discriminator was additionally trained to minimize \(L_{SL}\) and the generator was additionally trained to minimize \(L_{SA}\). In the original work, the authors set \(C=D_C\). For our work, we will be combining the Wasserstein and CAN methods. We used the following loss functions:
\[L_{disc}=\mathbb{E}_{x}[log(D(x)]+\mathbb{E}_{\mathcal{Z}}[log(1-D(G(\mathcal{Z}))]+L_{SL}\]
\[L_{gen}=-\mathbb{E}_{x}[log(D(x)]-\mathbb{E}_{\mathcal{Z}}[log(1-D(G(\mathcal{Z}))] +L_{SA}\]
\subsubsection{Diffusion}
A diffusion model aims to learn to iteratively remove the noise from a corrupted sample to restore the original. Starting with \(x_0\), the forward process \(q\) iteratively adds Gaussian noise to produce the noised version \(x_T\), using a noise schedule \(\beta_1...\beta_T\), which can be learned or manually set as a hyperparameter:
\[q(x_{1:T} | x_0)= \prod_{t=1}^{T} q(x_t | x_{t-1})\] 
\[q(x_t | x_{t-1}) =\mathcal{N} (x_t ; \sqrt{1-\beta_t} x_{t-1}, \beta_t \mathbf{I})\]
More importantly, we also want to model the reverse process \(p\), that turns a noisy sample \(x_T\) back into \(x_0\), conditioned on some context \(c\). As \(x_T\) is the fully noised version, \(p(x_T|c)=\mathcal{N}(x_T; \mathbf{0}, \mathbf{I})\)
\[p_{\theta}(x_{0:T} |c)=p(x_T|c) \prod_{t=1}^{T} p_{\theta} (x_{t-1} | x_t, c)\]
\[p_{\theta} (x_{t-1} | x_t,c)=\mathcal{N}(x_{t-1}; \mu_{\theta} (x_t,t,c), \Sigma_{\theta}(x_t,t,c))\]
We train \(\Sigma_{\theta}\) and \(\mu_{\theta}\) via optimizing the variational lower bound of the negative likelihood of the data:
\[\mathbb{E} [-\mathrm{log} p_{\theta} (x_0)]  \leq \mathbb{E} [-\mathrm{log} \frac{p_{\theta}(x_{0:T} |c)}{q(x_{1:T} | x_0)}=L\]
As shown by \citet{hodenoising2020}, this is equivalent to estimating the noise at each step using a model \(\epsilon_{\theta}\). So the loss to be optimized is:
\[L=\mathbb{E}_{x, \epsilon \sim \mathcal{N}(0,1), t} ||\epsilon - \epsilon_{\theta}(x_t,t)||_2^2\]

Once the model has been trained, the reverse process, aka inference, to generate a sample from noise \(x_T \sim \mathcal{N}(0,1)\) can be done iteratively by finding \(x_t-1\) given \(x_t, \alpha_t=1-\beta_t, \Bar{\alpha_t}=\prod_s^t \alpha_s, \mathcal{Z} \sim \mathcal{N}(0,1)\) and \(\sigma_t^2=\beta_t\) or \(\sigma_t^2=\frac{1-\alpha_{t-1}}{1-\alpha_t} \beta_t\) :
\[x_{t-1}=\frac{1}{\sqrt{\alpha_t}} (x_t-\frac{1-\alpha_t}{\sqrt{1-\Bar{\alpha_t}}} \epsilon_{\theta}(x_t,t) )+\sigma_t \mathcal{Z}\]
Acording to \citet{hodenoising2020}, both versions of \(\sigma_t\) had similar results. In our case, we used \(\sigma_t^2=\frac{1-\alpha_{t-1}}{1-\alpha_t} \beta_t\).

A Variational Autoencoder \citep{kingma2022autoencoding} consists of an encoder \(\mathcal{E}: \mathbb{R}^{h \times w \times 3} \rightarrow \mathbb{R}^{h_z \times w_z \times c_z} \) to map an image into a lower-dimensional latent space, and a decoder \(\mathcal{D}: \mathbb{R}^{h_z \times w_z \times c_z}  \rightarrow \mathbb{R}^{h \times w \times 3} \) to reverse this process. \citet{Rombach_2022_CVPR} performs diffusion but uses the latent representation of images \(z_0=\mathcal{E}(x_0)\):
\[L=\mathbb{E}_{x, \epsilon \sim \mathcal{N}(0,1), t} ||\epsilon - \epsilon_{\theta}(z_t,t)||_2^2\]
This method, which we employed in this work, is known as stable diffusion. The encoding and decoding between the image dimensions and the latent dimensions is often implicit, and for the rest of the paper we will use \(x_t\) not \(z_t\), as is common in the literature.

\subsubsection{Markov Decision Processes}
A Markov Decision Process \citep{bellman1957markovian} is defined as a tuple \((S, A, p_0, P, R)\) that models the actions of an agent in some environment with discrete time-steps. 
\begin{itemize}
    \item[] \(S\) is the state space, the set of states the environment can be in.
    \item[]  \(A\) is the set of actions that the agent can take. 
    \item[] \(p_0\) is the initial distributions of states \(s \in S\) when \(t=0\).
    \item[]  \(P_a(s, s')\) is the probability of transitioning from state \(s\) at time \(t\) to \(s'\) at \(t+1\) when the agent has taken action \(a \in A\).
    \item[]  The reward function \(R(s_t,a_t)\) returns a reward a time \(t\) given the action \(a_t\) the agent takes and the state of the environment \(s_t\).
\end{itemize}
The agents actions are determined by the policy \(\pi(a|s)\) that maps actions to states. The series of state-action pairs for each timestep is called a trajectory \(\tau=(s_0,a_0....s_T,a_T)\). Using policy-gradient as opposed to value-based RL, we train \(\pi\) by maximizing the reward \(R\) over the trajectories sampled from the policy:
\[\mathcal{J}_{RL}(\pi)=\mathbb{E}_{\tau \sim p(\tau | \pi)}[\sum_{t=0}^{T} R(s_t, a_t)]\]
\subsubsection{Denoising Diffusion Proximal Optimisation}
Introduced by \citet{black2023training}, Denoising Diffusion Proximal Optimisation, or DDPO, represents the Diffusion Process as a Markov Decision Process. A similar method was also pursued by \citet{fan2023dpok}.
\[a_t \overset{\Delta}{=} x_{t-1}\]
\[s_t \overset{\Delta}{=} (c,t,x_t) \]
\[  \pi(a_t | s_t) \overset{\Delta}{=} p_{\theta}(x_{t-1} | x_t, c)\]
\[p_0(s_0) \overset{\Delta}{=} (p(c), \delta_T, \mathcal{N}(0, \mathbf{I})) \]
\[ P(s_{t+1}) \overset{\Delta}{=} (\delta_c, \delta_{t-1}, \delta_{x_{t-1}}) \]
\[R(s_t, a_t) \overset{\Delta}{=} r(x_0,c) \]
\[\mathcal{J}_{RL}(\pi) \overset{\Delta}{=} \mathcal{J}_{DDRL}(\theta)= \mathbb{E}_{c \sim p(c), x_0 \sim p_{\theta}(x_0 | c)}[r(x_0,c)]\]

Reinforcement learning training was then applied to a pretrained diffusion model, which in our case was Stable Diffusion 2 \citep{Rombach_2022_CVPR}.  Following \citet{schulman2017proximal}, \citet{black2023training} also implemented clipping to protect the policy gradient \(\nabla_{\theta} \mathcal{J}_{DDRL}\) from excessively large updates. We largely follow their method but use a different reward function. We fine-tune off of the pre-existing \textbf{stabilityai/stable-diffusion-2-base} checkpoint \citep{Rombach_2022_CVPR} downloaded from \url{https://huggingface.co/stabilityai/stable-diffusion-2-base}.

\subsection{Reward Function}\label{reward}
In the original paper, the authors used four different reward functions for four different tasks. For example, they used a scorer trained on the LAION dataset \citep{laion} as the reward function to improve the aesthetic quality of generated outputs. In this paper, we use the reward model based on \citet{ElgammalLEM17}, where the model is rewarded for stylistic ambiguity. Given a generated image \(x_0 \in \mathbb{R}^{h \times w \times 3}\) and a classifier \(C: \mathbb{R}^{h \times w \times 3} \rightarrow \mathbb{R}^{N}\) we want to maximize:
\[R(x_0) = - \textbf{CE}(C(x_0),U) \]
where \(\textbf{CE}\) is the cross entropy.

\subsection{Data}
Starting with the WikiArt dataset \citep{wikiartSalehE15}, we used 1000 images from each class, oversampling when necessary, to balance the distributions between classes, to train the CAN. To train the diffusion model, we prompted the model  by concatenating a randomly selected medium prompt from (\textbf{painting of }, \textbf{picture of }, \textbf{drawing of }) to a randomly selected subject prompt (\textbf{a man}, \textbf{a woman}, \textbf{a landscape}, \textbf{nature}, \textbf{a building}, \textbf{an animal}, \textbf{shapes}, \textbf{ an object}). An example prompt would be \textbf{picture of an animal}. With 10\% probability we would set the prompt to the null string in order to train the model unconditionally as well.

\subsection{Choice of Classifier}
Style ambiguity loss relies on some classifier \(C\). We are exploring four versions of this classifier.

\subsubsection{DCGAN-Based Classifier}
We can use the classification module of the discriminator as the classifier in the reward function, setting \(C=D_C\). In the case of the CAN, \(D_C\) is trained jointly along with the generator. In the case of DDPO, we use a pretrained \(D_C\) from the CAN discriminator (which we call Diffusion DCGAN Based). 

\subsubsection{CLIP-Based Classifier}
Given text \(\in \R^{text}\) and an image \(\in \R^{h \times w \times 3}\),we can use a pretrained CLIP \citep{radford2021clip} model, that can return a similarity score for each image-text pair: \(CLIP: \R^{text} \times \R^{h \times w \times 3} \rightarrow \R \). CLIP is a multimodal foundation model trained using contrastive learning \citep{jaiswal2021survey} on a dataset of approximately 400 million text-image pairs. For each generated image \(x_0\), for each class name \(s_i, 1 \le i \le N_s\), we find \(CLIP(s_i,x_0)\). We can then create a vector \((CLIP(s_1,x_0), CLIP(s_2,x_0),,,CLIP(s_{N_s},x_0))\) and then use softmax to normalize the vector and define the result as \(C_{CLIP}(x_0)\). Formally:
\[C_{CLIP}(x_0)=\textbf{softmax}((CLIP(s_1,x_0), CLIP(s_2,x_0),,,CLIP(s_{N_s},x_0))\]
Then we set \(C=C_{CLIP}\). We discard the results of \(D_C\) when using a CLIP-Based Classifier with CAN. We used the 27 style classes in the WikArt dataset \citep{wikiartSalehE15} as \(s_i, 1 \le i \le N_s\). A list of said classes can be found in Appendix \ref{styleclasses}. We used the \textbf{clip-vit-large-patch14} CLIP checkpoint downloaded from \url{https://huggingface.co/openai/clip-vit-large-patch14}.
\subsubsection{K-Means Text and Image Based Classifiers}
Alternatively, when we have \(N_s\) text labels or \(N_I\) source images, we can embed the labels or images into the CLIP embedding space \(\in \R^{768}\) and perform k-means clustering to generate k centers. Given a CLIP Embedder \(E: \R^{h \times w \times 3} \rightarrow \R^{768}\) mapping images to embeddings, and the k centers \(c_1, c_2,,,c_k\) we can create a vector \((\frac{1}{||E(x_0)-c_1||},\frac{1}{||E(x_0)-c_2||},,,,\frac{1}{||E(x_0)-c_k||}\) and then use softmax to normalize the vector and define the result as \(C_{KMEANS}\). Formally:
\[C_{KMEANS}(x_0)=\textbf{softmax}(\frac{1}{||E(x_0)-c_1||},\frac{1}{||E(x_0)-c_2||},,,,\frac{1}{||E(x_0)-c_k||})\]
Then we set \(C=C_{KMEANS}\). We used two sets of centers: one set from performing k-means clustering on the WikiArt images, which we called \textbf{K Means Image Based}, and one set from clustering the names of the 27 style classes, which we call \textbf{K Means Text Based}.

\section{Results}

We generated all images with width and height = 512. The authors used width and height = 256 in the original CAN paper. However, given that larger, more detailed images are preferred by most people, we thought it more relevant to focus on larger images. Refer to appendix \ref{smaller} for results on smaller images and examples. Table \ref{tab:ddpo-eval} shows a few  DDPO images with the prompts used to generate them. Appendix \ref{images} shows more examples generated using different prompts. 

\exampleimgtable{512}{Example Images}{ddpo-eval}{22}

\subsection{Quantitative Evaluation}
We generated 100 images using the same prompts the models were trained on for each model.
We used three quantitative metrics to score the models
\begin{itemize}
    \item \textbf{AVA Score:} Consisting of CLIP+Multi-Layer Perceptron \citep{haykin2000neural}, the AVA model was trained on the AVA dataset \citep{avamurray} of images and average rankings by human subjects, in order to learn to approximate human preferences given an image. We used the CLIP model weights from the \textbf{clip-vit-large-patch14} checkpoint and the Multi-Layer Perceptron weights downloaded from \url{https://huggingface.co/trl-lib/ddpo-aesthetic-predictor}.
    \item \textbf{Image Reward:} The image reward model \citep{xu2023imagereward} was trained to score images given their text description based on a dataset of images and human rankings. We used the \textbf{image-reward} python library found at \url{https://github.com/THUDM/ImageReward/tree/main}.
    \item \textbf{Prompt Similarity:} Given the CLIP model's ability to embed images and text into the same space, we can measure the similarity between an image and its source prompt by finding the cosine similarity between the two CLIP embeddings. We used the \textbf{clip-vit-large-patch14} checkpoint.
\end{itemize}

\begin{table}[h]

\centering
\begin{tabular}{||c | c c c ||} 
\hline
Model & AVA Score & Image Reward & Prompt Similarity  \\
 \hline
 Diffusion- CLIP Based & 4.18 & -1.75 & 0.24 \\
Diffusion- K-Means Text Based & \textbf{4.60} & -1.21 & \textbf{0.26} \\
Diffusion- K-Means Image Based & 4.38 & \textbf{-0.90} & \textbf{0.26} \\
Diffusion- DCGAN Based  & 4.26 & -1.58 & \textbf{0.26} \\
\hline
\end{tabular}
\caption{Scores (Image Dim 512)}
\label{table:ava}
\end{table}

Results of our experiments are shown in table \ref{table:ava}. The best scores are bolded. There was little variance in prompt similarity. However, both K-Means-based approaches improved upon the DCGAN-based approach in terms of the two metrics for human preferences, showing that our method improves upon the past work aesthetically while also circumventing the costly training time of using a CAN or needing a labeled dataset for training the style classifier component of the CAN.

\subsection{Comparison with Baseline}
It is worth contrasting our trained DDPO model with the default pretrained \textbf{stabilityai/stable-diffusion-2-base} checkpoint diffusion model we are fine-tuning \citep{Rombach_2022_CVPR}. This allows us to better visualize the difference the DDPO training with style ambiguity loss makes. We assumed that the DDPO images might be similar to the baseline model images generated with fewer steps, so we compared DDPO Images to the baseline using 30,15, and 10 inference steps, as seen in figure \ref{tab:ddpo-vanilla}. 

\begin{center}
    \begin{table}[h]
        \centering
        \begin{tabular}{|c|c|c|c|c|c|c|c|}
        \hline
           Prompt  &  CLIP Based & \vbox{\hbox{\strut K Means} \hbox{\strut Text}} & \vbox{\hbox{\strut K Means} \hbox{\strut Image}} & DCGAN & \vbox{\hbox{\strut Baseline} \hbox{\strut (30 Steps)}} & \vbox{\hbox{\strut Baseline} \hbox{\strut (15 Steps)}} & \vbox{\hbox{\strut Baseline} \hbox{\strut (10 Steps)}} \\
        \hline
      \vbox{\hbox{\strut a painting} \hbox{\strut of a } \hbox{\strut landscape} \hbox{\strut }}
      & \includegraphics[scale=0.2]{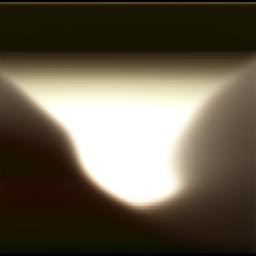}
      & \includegraphics[scale=0.2]{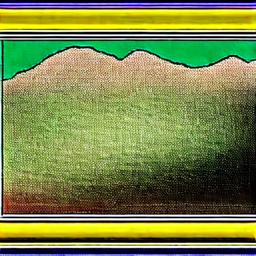}
      & \includegraphics[scale=0.2]{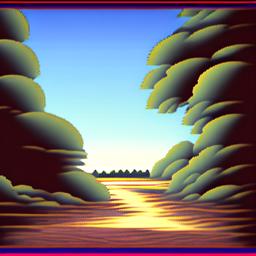}
      & \includegraphics[scale=0.2]{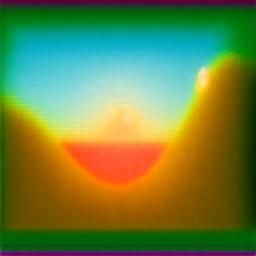}
      & \includegraphics[scale=0.2]{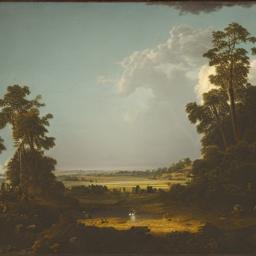}
      & \includegraphics[scale=0.2]{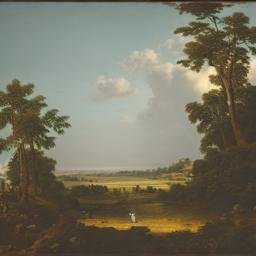}
      & \includegraphics[scale=0.2]{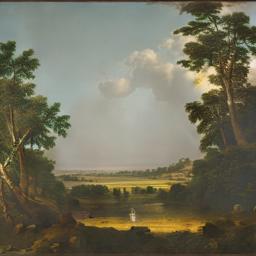} \\
      \vbox{\hbox{\strut a picture} \hbox{\strut of an } \hbox{\strut animal} \hbox{\strut }}
      & \includegraphics[scale=0.2]{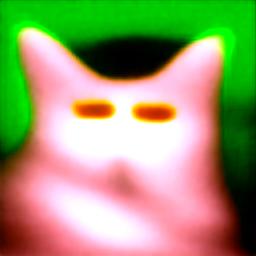}
      & \includegraphics[scale=0.2]{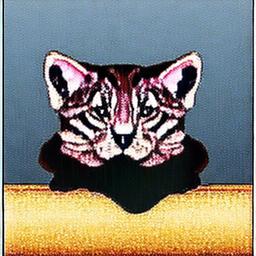}
      & \includegraphics[scale=0.2]{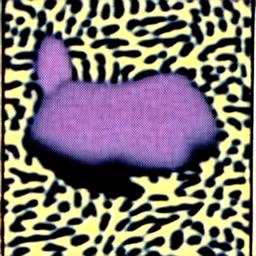}
      & \includegraphics[scale=0.2]{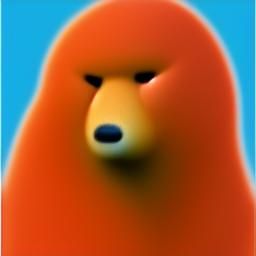}
      & \includegraphics[scale=0.2]{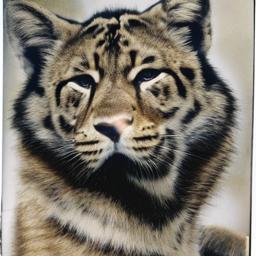}
      & \includegraphics[scale=0.2]{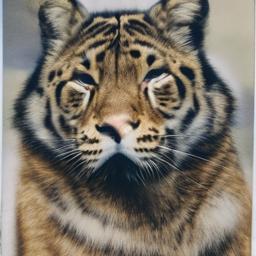}
      & \includegraphics[scale=0.2]{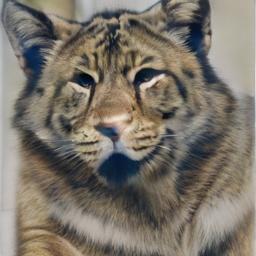} \\
      \vbox{\hbox{\strut a painting} \hbox{\strut of a } \hbox{\strut man} \hbox{\strut }}
      & \includegraphics[scale=0.2]{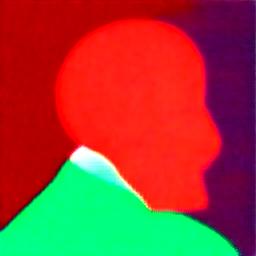}
      & \includegraphics[scale=0.2]{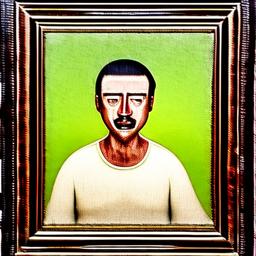}
      & \includegraphics[scale=0.2]{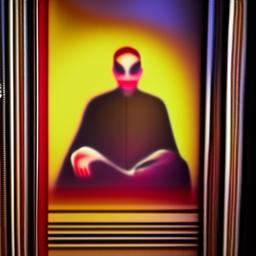}
      & \includegraphics[scale=0.2]{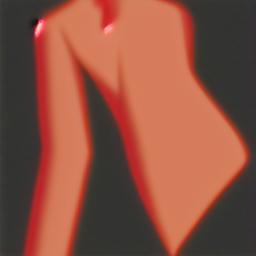}
      & \includegraphics[scale=0.2]{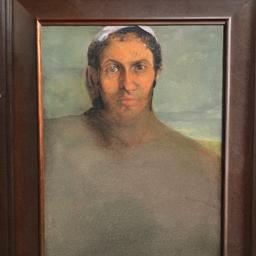}
      & \includegraphics[scale=0.2]{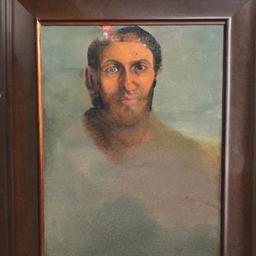}
      & \includegraphics[scale=0.2]{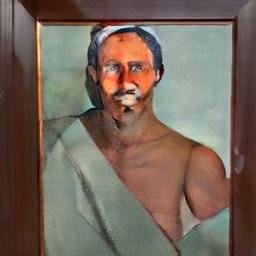} \\
      \hline
        \end{tabular}
        \caption{Example Images}
        \label{tab:ddpo-vanilla}
    \end{table}
\end{center}

In order to quantitatively compare our models to the baselines, we used the embedding of the \textbf{[CLS]} token from a vision transformer loaded from the \textbf{dino-vits16} checkpoint \citep{caron2021emerging} from \url{https://huggingface.co/facebook/dino-vits16}, as that encodes stylistic information \citep{tumanyan2022splicing,kwon2023diffusionbased}. We averaged the cosine similarity between style embeddings of each pair of images \((x,y_{30},y_{15},y_{10})\), where \(x\) was generated by the tuned model and \(y_{30},y_{15},y_{10}\) was generated by the baseline model using 30,15 and 10 inference steps respectively, using the same prompt and initial random seed. We did this 40 times. Average style cosine similarities between the DDPO-trained models and the baselines are shown in table \ref{table:style}. A lower style cosine similarity implies that the diffusion model has learned to successfully "deviate" from the baseline. All diffusion models model were more similar to the 30-step baseline, which implies that the diffusion models are not just learning to generate blurrier, less precise samples, but were learning a new "style" of art. 

\begin{table}[h!]
\centering
\begin{tabular}{| c |c | c | c   |} 
\hline
 & Baseline 30 Steps & Baseline 15 Steps & Baseline 10 Steps \\
 \hline
 Diffusion- CLIP Based & 0.29 & 0.27 & 0.23 \\
Diffusion- K-Means Text Based & 0.29 & 0.28 & 0.25 \\
Diffusion- K-Means Image Based & 0.30 & 0.30 & 0.29 \\
Diffusion- DCGAN Based &  0.26  & 0.23 & 0.21 \\
\hline
\end{tabular}
\caption{Style Similarities}
\label{table:style}
\end{table}
\pagebreak

\section{Conclusion}

Training models with stylistic ambiguity loss teaches them to be creative. This work introduces new forms of stylistic ambiguity loss that do not require training a classifier or GAN, which can be time-consuming and unstable \citep{saxena2023generative}. These new methods, particularly the K-Means-based approaches, scored higher than the traditional method on quantitative metrics of human judgement. Nonetheless, there are still more directions for this to go. Both the CLIP-based and K-Means Text-based style ambiguity losses require users to heuristically choose a set of styles to "deviate" from. In this work, we only used the 27 categories in the WikiArt dataset to be comparable to the original CAN paper. However, users may instead prefer a different set of styles or words, which may produce better or more interesting results. Additionally, the K-Means Image-based style ambiguity loss does not require a multimodal model like CLIP. We could have used any pretrained model to embed images into a lower-dimensional manifold, or trained a new one. Ergo, the K-means technique could be used for any medium, such as music \citep{patent_Elgammal_2022, audio_zhang2023survey}, new proteins \citep{winnifrith2023generative}, stories \citep{mori2022computational} and videos \citep{cho2024sora}.

\subsubsection*{Broader Impact Statement}
Many are concerned about the impacts of generative AI. By making art, this work infringes upon a domain once exclusive to humans. Companies have faced scrutiny for possibly using AI \citep{wizards}, and many creatives, such as screenwriters and actors, have voiced concerns about whether their jobs are safe \citep{npr}. Nonetheless, using AI can help humans by making them more efficient, providing inspiration, and generating ideas \citep{nyu, cornell, sciencefocus}. It's also not certain how copyright protection will function for AI-generated art \citep{copyright2023}, given copyright law is based on the premise that creative works originate solely from human authorship. Clear, consistent policies, both at the government level and by industry and/or academic groups, will be needed to mitigate the harm and maximize the benefits for all members of society. 

\section{Assistance}

\subsection*{Author Contributions}
This work was done without any outside assistance or collaboration.

\subsection*{Acknowledgements}
Rutgers Office of Advanced Research Computing kindly provided the computing infrastructure to run the experiments. A special thanks goes out to Dr. Ahmed Elgammal and Dr. Eugene White for their past advice prior to this work.

\bibliography{main}

\begin{thebibliography}{66}
\providecommand{\natexlab}[1]{#1}
\providecommand{\url}[1]{\texttt{#1}}
\expandafter\ifx\csname urlstyle\endcsname\relax
  \providecommand{\doi}[1]{doi: #1}\else
  \providecommand{\doi}{doi: \begingroup \urlstyle{rm}\Url}\fi

\bibitem[Arjovsky et~al.(2017)Arjovsky, Chintala, and Bottou]{arjovsky2017wasserstein}
Martin Arjovsky, Soumith Chintala, and Léon Bottou.
\newblock Wasserstein gan, 2017.

\bibitem[Bellman(1957)]{bellman1957markovian}
Richard Bellman.
\newblock A markovian decision process.
\newblock \emph{Journal of Mathematics and Mechanics}, 6\penalty0 (5):\penalty0 679--684, 1957.
\newblock URL \url{http://www.jstor.org/stable/24900506}.

\bibitem[Betker et~al.()Betker, Goh, Jing, TimBrooks, Wang, Li, LongOuyang, JuntangZhuang, JoyceLee, YufeiGuo, WesamManassra, PrafullaDhariwal, CaseyChu, YunxinJiao, and Ramesh]{BetkerImprovingIG}
James Betker, Gabriel Goh, Li~Jing, † TimBrooks, Jianfeng Wang, Linjie Li, † LongOuyang, † JuntangZhuang, † JoyceLee, † YufeiGuo, † WesamManassra, † PrafullaDhariwal, † CaseyChu, † YunxinJiao, and Aditya Ramesh.
\newblock Improving image generation with better captions.
\newblock URL \url{https://api.semanticscholar.org/CorpusID:264403242}.

\bibitem[Black et~al.(2023)Black, Janner, Du, Kostrikov, and Levine]{black2023training}
Kevin Black, Michael Janner, Yilun Du, Ilya Kostrikov, and Sergey Levine.
\newblock Training diffusion models with reinforcement learning, 2023.

\bibitem[Boden(1990)]{Boden}
Margaret Boden.
\newblock \emph{The Creative Mind}.
\newblock Abacus, 1990.

\bibitem[Campitiello(2023)]{cornell}
Jess Campitiello.
\newblock Ai vs. artist: The future of creativity, 2023.
\newblock URL \url{https://tech.cornell.edu/news/ai-vs-artist-the-future-of-creativity/}.

\bibitem[Caron et~al.(2021)Caron, Touvron, Misra, J\'egou, Mairal, Bojanowski, and Joulin]{caron2021emerging}
Mathilde Caron, Hugo Touvron, Ishan Misra, Herv\'e J\'egou, Julien Mairal, Piotr Bojanowski, and Armand Joulin.
\newblock Emerging properties in self-supervised vision transformers.
\newblock In \emph{Proceedings of the International Conference on Computer Vision (ICCV)}, 2021.

\bibitem[Cho et~al.(2024)Cho, Puspitasari, Zheng, Zheng, Lee, Kim, Hong, and Zhang]{cho2024sora}
Joseph Cho, Fachrina~Dewi Puspitasari, Sheng Zheng, Jingyao Zheng, Lik-Hang Lee, Tae-Ho Kim, Choong~Seon Hong, and Chaoning Zhang.
\newblock Sora as an agi world model? a complete survey on text-to-video generation, 2024.

\bibitem[Cropley(2006)]{cropley}
Arthur Cropley.
\newblock In praise of convergent thinking.
\newblock \emph{Creativity Research Journal}, 18\penalty0 (3):\penalty0 391--404, 2006.
\newblock \doi{10.1207/s15326934crj1803\_13}.
\newblock URL \url{https://doi.org/10.1207/s15326934crj1803_13}.

\bibitem[Darling(2022)]{sciencefocus}
Kate Darling.
\newblock Ai image generators will help artists, not replace them, 2022.
\newblock URL \url{https://www.sciencefocus.com/news/ai-image-generators-will-help-artists-not-replace-them}.

\bibitem[del Barco(2023)]{npr}
Mandalit del Barco.
\newblock Some sag-aftra members are concerned about ai provisions in tentative deal, 2023.
\newblock URL \url{https://www.npr.org/2023/11/30/1216005659/some-sag-aftra-members-are-concerned-about-ai-provisions-in-tentative-deal}.

\bibitem[DiPaola \& Gabora(2008)DiPaola and Gabora]{DiPaola_2008}
Steve DiPaola and Liane Gabora.
\newblock Incorporating characteristics of human creativity into an evolutionary art algorithm.
\newblock \emph{Genetic Programming and Evolvable Machines}, 10\penalty0 (2):\penalty0 97–110, December 2008.
\newblock ISSN 1573-7632.
\newblock \doi{10.1007/s10710-008-9074-x}.
\newblock URL \url{http://dx.doi.org/10.1007/s10710-008-9074-x}.

\bibitem[Dubey et~al.(2022)Dubey, Singh, and Chaudhuri]{dubey2022activation}
Shiv~Ram Dubey, Satish~Kumar Singh, and Bidyut~Baran Chaudhuri.
\newblock Activation functions in deep learning: A comprehensive survey and benchmark, 2022.

\bibitem[Dumoulin \& Visin(2018)Dumoulin and Visin]{dumoulin2018guide}
Vincent Dumoulin and Francesco Visin.
\newblock A guide to convolution arithmetic for deep learning, 2018.

\bibitem[Elgammal(2022)]{patent_Elgammal_2022}
Ahmed Elgammal.
\newblock Creative gan generating music deviating from style norms, Feb 2022.

\bibitem[Elgammal et~al.(2017)Elgammal, Liu, Elhoseiny, and Mazzone]{ElgammalLEM17}
Ahmed~M. Elgammal, Bingchen Liu, Mohamed Elhoseiny, and Marian Mazzone.
\newblock {CAN:} creative adversarial networks, generating "art" by learning about styles and deviating from style norms.
\newblock \emph{CoRR}, abs/1706.07068, 2017.
\newblock URL \url{http://arxiv.org/abs/1706.07068}.

\bibitem[Fan et~al.(2023)Fan, Watkins, Du, Liu, Ryu, Boutilier, Abbeel, Ghavamzadeh, Lee, and Lee]{fan2023dpok}
Ying Fan, Olivia Watkins, Yuqing Du, Hao Liu, Moonkyung Ryu, Craig Boutilier, Pieter Abbeel, Mohammad Ghavamzadeh, Kangwook Lee, and Kimin Lee.
\newblock Dpok: Reinforcement learning for fine-tuning text-to-image diffusion models, 2023.

\bibitem[Fortino(2023)]{nyu}
Andres Fortino.
\newblock Embracing creativity: How ai can enhance the creative process, 2023.
\newblock URL \url{https://www.sps.nyu.edu/homepage/emerging-technologies-collaborative/blog/2023/embracing-creativity-how-ai-can-enhance-the-creative-process.html}.

\bibitem[Goodfellow et~al.(2014)Goodfellow, Pouget-Abadie, Mirza, Xu, Warde-Farley, Ozair, Courville, and Bengio]{goodfellow2014generative}
Ian~J. Goodfellow, Jean Pouget-Abadie, Mehdi Mirza, Bing Xu, David Warde-Farley, Sherjil Ozair, Aaron Courville, and Yoshua Bengio.
\newblock Generative adversarial networks, 2014.

\bibitem[Gugger et~al.(2022)Gugger, Debut, Wolf, Schmid, Mueller, Mangrulkar, Sun, and Bossan]{accelerate2022}
Sylvain Gugger, Lysandre Debut, Thomas Wolf, Philipp Schmid, Zachary Mueller, Sourab Mangrulkar, Marc Sun, and Benjamin Bossan.
\newblock Accelerate: Training and inference at scale made simple, efficient and adaptable.
\newblock \url{https://github.com/huggingface/accelerate}, 2022.

\bibitem[Gutierrez(2024)]{wizards}
Luis~Joshua Gutierrez.
\newblock Wizards of the coast repeats anti-ai stance, fights accusation against latest magic the gathering promo, 2024.
\newblock URL \url{https://www.gamespot.com/articles/wizards-of-the-coast-repeats-anti-ai-stance-fights-accusation-against-latest-magic-the-gathering-promo/1100-6520153/}.

\bibitem[Haarnoja et~al.(2018)Haarnoja, Zhou, Abbeel, and Levine]{haarnoja2018soft}
Tuomas Haarnoja, Aurick Zhou, Pieter Abbeel, and Sergey Levine.
\newblock Soft actor-critic: Off-policy maximum entropy deep reinforcement learning with a stochastic actor, 2018.

\bibitem[Haykin(2000)]{haykin2000neural}
Simon Haykin.
\newblock Neural networks: A guided tour.
\newblock 2000.

\bibitem[Ho et~al.(2020)Ho, Jain, and Abbeel]{hodenoising2020}
Jonathan Ho, Ajay Jain, and Pieter Abbeel.
\newblock Denoising diffusion probabilistic models.
\newblock \emph{CoRR}, abs/2006.11239, 2020.
\newblock URL \url{https://arxiv.org/abs/2006.11239}.

\bibitem[Ioffe \& Szegedy(2015)Ioffe and Szegedy]{ioffe2015batch}
Sergey Ioffe and Christian Szegedy.
\newblock Batch normalization: Accelerating deep network training by reducing internal covariate shift, 2015.

\bibitem[Iqbal et~al.(2016)Iqbal, Guid, Colton, Krivec, Azman, and Haghighi]{iqbal2016digital}
Azlan Iqbal, Matej Guid, Simon Colton, Jana Krivec, Shazril Azman, and Boshra Haghighi.
\newblock The digital synaptic neural substrate: A new approach to computational creativity, 2016.

\bibitem[Jaiswal et~al.(2021)Jaiswal, Babu, Zadeh, Banerjee, and Makedon]{jaiswal2021survey}
Ashish Jaiswal, Ashwin~Ramesh Babu, Mohammad~Zaki Zadeh, Debapriya Banerjee, and Fillia Makedon.
\newblock A survey on contrastive self-supervised learning, 2021.

\bibitem[Karras et~al.(2018)Karras, Aila, Laine, and Lehtinen]{karras2018progressive}
Tero Karras, Timo Aila, Samuli Laine, and Jaakko Lehtinen.
\newblock Progressive growing of gans for improved quality, stability, and variation, 2018.

\bibitem[Karras et~al.(2019)Karras, Laine, and Aila]{karras2019stylebased}
Tero Karras, Samuli Laine, and Timo Aila.
\newblock A style-based generator architecture for generative adversarial networks, 2019.

\bibitem[Kingma \& Welling(2022)Kingma and Welling]{kingma2022autoencoding}
Diederik~P Kingma and Max Welling.
\newblock Auto-encoding variational bayes, 2022.

\bibitem[Kiran et~al.(2021)Kiran, Sobh, Talpaert, Mannion, Sallab, Yogamani, and Pérez]{kiran2021deep}
B~Ravi Kiran, Ibrahim Sobh, Victor Talpaert, Patrick Mannion, Ahmad A.~Al Sallab, Senthil Yogamani, and Patrick Pérez.
\newblock Deep reinforcement learning for autonomous driving: A survey, 2021.

\bibitem[Kozodoi(2021)]{kozodoi2021gradient}
N~Kozodoi.
\newblock Gradient accumulation in pytorch, 2021.

\bibitem[Kwon \& Ye(2023)Kwon and Ye]{kwon2023diffusionbased}
Gihyun Kwon and Jong~Chul Ye.
\newblock Diffusion-based image translation using disentangled style and content representation, 2023.

\bibitem[Lacoste et~al.(2019)Lacoste, Luccioni, Schmidt, and Dandres]{lacoste2019quantifying}
Alexandre Lacoste, Alexandra Luccioni, Victor Schmidt, and Thomas Dandres.
\newblock Quantifying the carbon emissions of machine learning, 2019.

\bibitem[Lillicrap et~al.(2019)Lillicrap, Hunt, Pritzel, Heess, Erez, Tassa, Silver, and Wierstra]{lillicrap2019ddpg}
Timothy~P. Lillicrap, Jonathan~J. Hunt, Alexander Pritzel, Nicolas Heess, Tom Erez, Yuval Tassa, David Silver, and Daan Wierstra.
\newblock Continuous control with deep reinforcement learning, 2019.

\bibitem[Luo(2022)]{luo2022understanding}
Calvin Luo.
\newblock Understanding diffusion models: A unified perspective, 2022.

\bibitem[Lutz et~al.(2023)Lutz, Wang, Norn, Courbet, Borst, Zhao, Dosey, Cao, Xu, Leaf, et~al.]{lutz2023top}
Isaac~D Lutz, Shunzhi Wang, Christoffer Norn, Alexis Courbet, Andrew~J Borst, Yan~Ting Zhao, Annie Dosey, Longxing Cao, Jinwei Xu, Elizabeth~M Leaf, et~al.
\newblock Top-down design of protein architectures with reinforcement learning.
\newblock \emph{Science}, 380\penalty0 (6642):\penalty0 266--273, 2023.

\bibitem[Maas et~al.(2013)Maas, Hannun, Ng, et~al.]{maas2013rectifier}
Andrew~L Maas, Awni~Y Hannun, Andrew~Y Ng, et~al.
\newblock Rectifier nonlinearities improve neural network acoustic models.
\newblock In \emph{Proc. icml}, volume~30, pp.\ ~3. Atlanta, GA, 2013.

\bibitem[Mangrulkar et~al.(2022)Mangrulkar, Gugger, Debut, Belkada, Paul, and Bossan]{peft2022}
Sourab Mangrulkar, Sylvain Gugger, Lysandre Debut, Younes Belkada, Sayak Paul, and Benjamin Bossan.
\newblock Peft: State-of-the-art parameter-efficient fine-tuning methods.
\newblock \url{https://github.com/huggingface/peft}, 2022.

\bibitem[McCorduck(1991)]{mccorduck1991aaron}
P.~McCorduck.
\newblock \emph{Aaron's Code: Meta-art, Artificial Intelligence, and the Work of Harold Cohen}.
\newblock W.H. Freeman, 1991.
\newblock ISBN 9780716721734.
\newblock URL \url{https://books.google.com/books?id=r3UyBgAAQBAJ}.

\bibitem[Mori et~al.(2022)Mori, Yamane, Mukuta, and Harada]{mori2022computational}
Yusuke Mori, Hiroaki Yamane, Yusuke Mukuta, and Tatsuya Harada.
\newblock Computational storytelling and emotions: A survey, 2022.

\bibitem[Murray et~al.(2016)Murray, Marchesotti, and Perronnin]{avamurray}
Naila Murray, Luca Marchesotti, and Florent Perronnin.
\newblock Ava: A large-scale database for aesthetic visual analysis.
\newblock 2016.
\newblock URL \url{https://github.com/imfing/ava_downloader}.

\bibitem[Paszke et~al.(2017)Paszke, Gross, Chintala, Chanan, Yang, DeVito, Lin, Desmaison, Antiga, and Lerer]{paszke2017pytorch}
Adam Paszke, Sam Gross, Soumith Chintala, Gregory Chanan, Edward Yang, Zachary DeVito, Zeming Lin, Alban Desmaison, Luca Antiga, and Adam Lerer.
\newblock Automatic differentiation in pytorch.
\newblock 2017.

\bibitem[Pawelec(2022)]{democracypawelec2022}
Maria Pawelec.
\newblock Deepfakes and democracy (theory): How synthetic audio-visual media for disinformation and hate speech threaten core democratic functions.
\newblock \emph{Digital Society}, 1, 09 2022.
\newblock \doi{10.1007/s44206-022-00010-6}.

\bibitem[Pedregosa et~al.(2011)Pedregosa, Varoquaux, Gramfort, Michel, Thirion, Grisel, Blondel, Prettenhofer, Weiss, Dubourg, Vanderplas, Passos, Cournapeau, Brucher, Perrot, and {{\'E}}douard Duchesnay]{sklearn:pedregosa11a}
Fabian Pedregosa, Ga{{\"e}}l Varoquaux, Alexandre Gramfort, Vincent Michel, Bertrand Thirion, Olivier Grisel, Mathieu Blondel, Peter Prettenhofer, Ron Weiss, Vincent Dubourg, Jake Vanderplas, Alexandre Passos, David Cournapeau, Matthieu Brucher, Matthieu Perrot, and {{\'E}}douard Duchesnay.
\newblock Scikit-learn: Machine learning in python.
\newblock \emph{Journal of Machine Learning Research}, 12\penalty0 (85):\penalty0 2825--2830, 2011.
\newblock URL \url{http://jmlr.org/papers/v12/pedregosa11a.html}.

\bibitem[Qiang \& Zhongli(2011)Qiang and Zhongli]{qiang2011rl}
Wang Qiang and Zhan Zhongli.
\newblock Reinforcement learning model, algorithms and its application.
\newblock In \emph{2011 International Conference on Mechatronic Science, Electric Engineering and Computer (MEC)}, pp.\  1143--1146, 2011.
\newblock \doi{10.1109/MEC.2011.6025669}.

\bibitem[Radford et~al.(2016)Radford, Metz, and Chintala]{radford2016unsupervised}
Alec Radford, Luke Metz, and Soumith Chintala.
\newblock Unsupervised representation learning with deep convolutional generative adversarial networks, 2016.

\bibitem[Radford et~al.(2021)Radford, Kim, Hallacy, Ramesh, Goh, Agarwal, Sastry, Askell, Mishkin, Clark, Krueger, and Sutskever]{radford2021clip}
Alec Radford, Jong~Wook Kim, Chris Hallacy, Aditya Ramesh, Gabriel Goh, Sandhini Agarwal, Girish Sastry, Amanda Askell, Pamela Mishkin, Jack Clark, Gretchen Krueger, and Ilya Sutskever.
\newblock Learning transferable visual models from natural language supervision, 2021.

\bibitem[Rombach et~al.(2022)Rombach, Blattmann, Lorenz, Esser, and Ommer]{Rombach_2022_CVPR}
Robin Rombach, Andreas Blattmann, Dominik Lorenz, Patrick Esser, and Bj\"orn Ommer.
\newblock High-resolution image synthesis with latent diffusion models.
\newblock In \emph{Proceedings of the IEEE/CVF Conference on Computer Vision and Pattern Recognition (CVPR)}, pp.\  10684--10695, June 2022.

\bibitem[Saharia et~al.(2022)Saharia, Chan, Saxena, Li, Whang, Denton, Ghasemipour, Ayan, Mahdavi, Lopes, Salimans, Ho, Fleet, and Norouzi]{saharia2022photorealistic}
Chitwan Saharia, William Chan, Saurabh Saxena, Lala Li, Jay Whang, Emily Denton, Seyed Kamyar~Seyed Ghasemipour, Burcu~Karagol Ayan, S.~Sara Mahdavi, Rapha~Gontijo Lopes, Tim Salimans, Jonathan Ho, David~J Fleet, and Mohammad Norouzi.
\newblock Photorealistic text-to-image diffusion models with deep language understanding, 2022.

\bibitem[Sahu et~al.(2023)Sahu, Mokhade, and Bokde]{sahu2023overview}
Santosh~Kumar Sahu, Anil Mokhade, and Neeraj~Dhanraj Bokde.
\newblock An overview of machine learning, deep learning, and reinforcement learning-based techniques in quantitative finance: Recent progress and challenges.
\newblock \emph{Applied Sciences}, 13\penalty0 (3):\penalty0 1956, 2023.

\bibitem[Saleh \& Elgammal(2015)Saleh and Elgammal]{wikiartSalehE15}
Babak Saleh and Ahmed~M. Elgammal.
\newblock Large-scale classification of fine-art paintings: Learning the right metric on the right feature.
\newblock \emph{CoRR}, abs/1505.00855, 2015.
\newblock URL \url{http://arxiv.org/abs/1505.00855}.

\bibitem[Saxena \& Cao(2023)Saxena and Cao]{saxena2023generative}
Divya Saxena and Jiannong Cao.
\newblock Generative adversarial networks (gans survey): Challenges, solutions, and future directions, 2023.

\bibitem[Schuhmann \& Beaumont(2022)Schuhmann and Beaumont]{laion}
Christoph Schuhmann and Romain Beaumont, 2022.
\newblock URL \url{https://laion.ai/blog/laion-aesthetics/}.

\bibitem[Schulman et~al.(2017{\natexlab{a}})Schulman, Levine, Moritz, Jordan, and Abbeel]{schulman2017trust}
John Schulman, Sergey Levine, Philipp Moritz, Michael~I. Jordan, and Pieter Abbeel.
\newblock Trust region policy optimization, 2017{\natexlab{a}}.

\bibitem[Schulman et~al.(2017{\natexlab{b}})Schulman, Wolski, Dhariwal, Radford, and Klimov]{schulman2017proximal}
John Schulman, Filip Wolski, Prafulla Dhariwal, Alec Radford, and Oleg Klimov.
\newblock Proximal policy optimization algorithms, 2017{\natexlab{b}}.

\bibitem[Silver et~al.(2017)Silver, Schrittwieser, Simonyan, Antonoglou, Huang, Guez, Hubert, baker, Lai, Bolton, Chen, Lillicrap, Hui, Sifre, van~den Driessche, Graepel, and Hassabis]{Silver2017MasteringTG}
David Silver, Julian Schrittwieser, Karen Simonyan, Ioannis Antonoglou, Aja Huang, Arthur Guez, Thomas Hubert, Lucas baker, Matthew Lai, Adrian Bolton, Yutian Chen, Timothy~P. Lillicrap, Fan Hui, L.~Sifre, George van~den Driessche, Thore Graepel, and Demis Hassabis.
\newblock Mastering the game of go without human knowledge.
\newblock \emph{Nature}, 550:\penalty0 354--359, 2017.
\newblock URL \url{https://api.semanticscholar.org/CorpusID:205261034}.

\bibitem[Sohl{-}Dickstein et~al.(2015)Sohl{-}Dickstein, Weiss, Maheswaranathan, and Ganguli]{Sohl-DicksteinW15}
Jascha Sohl{-}Dickstein, Eric~A. Weiss, Niru Maheswaranathan, and Surya Ganguli.
\newblock Deep unsupervised learning using nonequilibrium thermodynamics.
\newblock \emph{CoRR}, abs/1503.03585, 2015.
\newblock URL \url{http://arxiv.org/abs/1503.03585}.

\bibitem[Sutton et~al.(1999)Sutton, McAllester, Singh, and Mansour]{sutton1999policygradient}
Richard~S Sutton, David McAllester, Satinder Singh, and Yishay Mansour.
\newblock Policy gradient methods for reinforcement learning with function approximation.
\newblock In S.~Solla, T.~Leen, and K.~M\"{u}ller (eds.), \emph{Advances in Neural Information Processing Systems}, volume~12. MIT Press, 1999.
\newblock URL \url{https://proceedings.neurips.cc/paper_files/paper/1999/file/464d828b85b0bed98e80ade0a5c43b0f-Paper.pdf}.

\bibitem[Tumanyan et~al.(2022)Tumanyan, Bar-Tal, Bagon, and Dekel]{tumanyan2022splicing}
Narek Tumanyan, Omer Bar-Tal, Shai Bagon, and Tali Dekel.
\newblock Splicing vit features for semantic appearance transfer, 2022.

\bibitem[von Platen et~al.(2022)von Platen, Patil, Lozhkov, Cuenca, Lambert, Rasul, Davaadorj, Nair, Paul, Berman, Xu, Liu, and Wolf]{von-platen-etal-2022-diffusers}
Patrick von Platen, Suraj Patil, Anton Lozhkov, Pedro Cuenca, Nathan Lambert, Kashif Rasul, Mishig Davaadorj, Dhruv Nair, Sayak Paul, William Berman, Yiyi Xu, Steven Liu, and Thomas Wolf.
\newblock Diffusers: State-of-the-art diffusion models.
\newblock \url{https://github.com/huggingface/diffusers}, 2022.

\bibitem[von Werra et~al.(2020)von Werra, Belkada, Tunstall, Beeching, Thrush, Lambert, and Huang]{vonwerra2022trl}
Leandro von Werra, Younes Belkada, Lewis Tunstall, Edward Beeching, Tristan Thrush, Nathan Lambert, and Shengyi Huang.
\newblock Trl: Transformer reinforcement learning.
\newblock \url{https://github.com/huggingface/trl}, 2020.

\bibitem[Watiktinnakorn et~al.(2023)Watiktinnakorn, Seesai, and Kerdvibulvech]{copyright2023}
Chawinthorn Watiktinnakorn, Jirawat Seesai, and Chutisant Kerdvibulvech.
\newblock Blurring the lines: how ai is redefining artistic ownership and copyright.
\newblock \emph{Discover Artificial Intelligence}, 3, 11 2023.
\newblock \doi{10.1007/s44163-023-00088-y}.

\bibitem[Winnifrith et~al.(2023)Winnifrith, Outeiral, and Hie]{winnifrith2023generative}
Adam Winnifrith, Carlos Outeiral, and Brian Hie.
\newblock Generative artificial intelligence for de novo protein design, 2023.

\bibitem[Xu et~al.(2023)Xu, Liu, Wu, Tong, Li, Ding, Tang, and Dong]{xu2023imagereward}
Jiazheng Xu, Xiao Liu, Yuchen Wu, Yuxuan Tong, Qinkai Li, Ming Ding, Jie Tang, and Yuxiao Dong.
\newblock Imagereward: Learning and evaluating human preferences for text-to-image generation, 2023.

\bibitem[Zhang et~al.(2023)Zhang, Zhang, Zheng, Zhang, Qamar, Bae, and Kweon]{audio_zhang2023survey}
Chenshuang Zhang, Chaoning Zhang, Sheng Zheng, Mengchun Zhang, Maryam Qamar, Sung-Ho Bae, and In~So Kweon.
\newblock A survey on audio diffusion models: Text to speech synthesis and enhancement in generative ai, 2023.

\end{thebibliography}
\bibliographystyle{tmlr}
\appendix

\section{WikiArt Style Classes}\label{styleclasses}
The 27 WikiArt style classes are listed in table \ref{tab:styles}
\begin{table}[h]
\centering
\begin{tabular}{|c | c | c|} 
 \hline
contemporary-realism & art-nouveau-modern & abstract-expressionism \\ 
\hline
northern-renaissance & mannerism-late-renaissance & early-renaissance \\
\hline
realism &
action-painting &
color-field-painting \\
\hline
pop-art & new-realism & pointillism\\
\hline
expressionism &
analytical-cubism &
symbolism \\
\hline
fauvism &
minimalism &
cubism  \\
\hline
romanticism &
ukiyo-e &
high-renaissance \\
\hline
synthetic-cubism &
baroque &
post-impressionism \\
\hline
impressionism &
rococo &
na-ve-art-primitivism \\
 \hline
\end{tabular}
\caption{Styles}
\label{tab:styles}
\end{table}

\section{Images}\label{images}
Additional text prompts and the corresponding generated images using the DDPO models can be seen in Figures \ref{tab:ddpo-example-1} and \ref{tab:ddpo-example-2}. 

\exampleimgtable{512}{Example Images (512)}{ddpo-example-1}{19}
\exampleimgtable{512}{Example Images (512)}{ddpo-example-2}{20}

\section{Lower Dimensional Images}\label{smaller}
All experiments and images portrayed were done using images of dimension 512. However, we also repeated the experiments using smaller images. We briefly illustrate some example images in tables \ref{tab:ddpo-eval-256}, \ref{tab:ddpo-eval-128} and \ref{tab:ddpo-eval-64} as well as quantitative evaluations in tables \ref{table:ava256}, \ref{table:ava128} and \ref{table:ava64}.

\exampleimgtable{256}{Example Images (256)}{ddpo-eval-256}{19}

\exampleimgtable{128}{Example Images (128)}{ddpo-eval-128}{19}

\exampleimgtable{64}{Example Images (64)}{ddpo-eval-64}{19}
\begin{table}[h]

\centering
\begin{tabular}{||c | c c c ||} 
\hline
Model & AVA Score & Image Reward & Prompt Alignment \\
 \hline
 Diffusion- CLIP Based & 4.83 & -0.42 & 0.27 \\
Diffusion- K-Means Text Based & 4.91 & -0.64 & 0.27 \\
Diffusion- K-Means Image Based & 4.91 & -0.44 & 0.28 \\
Diffusion- DCGAN Based  & 4.95 & -0.33 & 0.27 \\
\hline

\end{tabular}

\label{table:ava256}
\caption{Scores (256)}
\end{table}

\begin{table}[h]
\centering
\begin{tabular}{||c | c c c ||} 
\hline
Model & AVA Score & Image Reward & Prompt Alignment \\
 \hline
 Diffusion- CLIP Based & 4.47 & -0.74 & 0.27 \\
Diffusion- K-Means Text Based & 4.77 & -0.37 & 0.28 \\
Diffusion- K-Means Image Based & 3.78 & -1.36 & 0.25 \\
Diffusion- DCGAN Based  & 4.66 & -0.14 & 0.29 \\
\hline

\end{tabular}

\label{table:ava128}
\caption{Scores (128)}
\end{table}

\begin{table}[h]
\centering
\begin{tabular}{||c | c c c ||} 
\hline
Model & AVA Score & Image Reward & Prompt Alignment \\
 \hline
 Diffusion- CLIP Based & 4.21 & -1.44 & 0.26 \\
Diffusion- K-Means Text Based & 4.36 & -1.00 &  0.27 \\
Diffusion- K-Means Image Based & 4.38 & -1.55 & 0.26 \\
Diffusion- DCGAN Based  & 4.62 & -0.22 & 0.28 \\
\hline

\end{tabular}

\label{table:ava64}
\caption{Scores (64)}

\end{table}

\section{Training}\label{hyper}

For reproducibility and transparency, the hyperparameters are listed in table \ref{tab:ddpo-hyper} and table \ref{tab:can-hyper}. All experiments were implemented in Python, building the models in \textbf{pytorch} \citep{paszke2017pytorch} using \textbf{accelerate} \citep{accelerate2022} for efficient training. The diffusion models also relied on the \textbf{trl} \citep{vonwerra2022trl}, \textbf{diffusers} \citep{von-platen-etal-2022-diffusers} and \textbf{peft} \citep{peft2022} libraries. The K-Means clustering was done using the k means implementation from \textbf{scikit-learn} \citep{sklearn:pedregosa11a}.  A repository containing all code can be found on github at \url{https://github.com/jamesBaker361/clipcreate/tree/main}. Each experiment was run using two NVIDIA A100 GPUs with 40 GB RAM. Training times and estimated carbon emissions \citep{lacoste2019quantifying} calculated with \url{https://mlco2.github.io/impact#compute} are shown in table \ref{tab:training}. 

\begin{table}[h!]
\centering
\begin{tabular}{||c c||} 
 \hline
 Hyperparameter & Value  \\ [0.5ex] 
 \hline\hline
 Epochs & 50 \\ 
 Effective Batch Size & 8 \\
 Batches per Epoch & 32 \\
 Inference Steps per Image & 30 \\
 LORA Matrix Rank & 4 \\
 LORA \(\alpha\) & 4 \\
 Optimizer & AdamW  \\
 Learning Rate & 3e-4 \\
 AdamW \(\beta_1\) & 0.9 \\
 AdamW \(\beta_2\) & 0.99 \\
 AdamW Weight decay & 1e-4 \\
 AdamW \(\epsilon\) & 1e-8 \\
 \hline
\end{tabular}
\caption{DDPO Hyperparameters}
\label{tab:ddpo-hyper}
\end{table}

\subsection{Batch Size}
For all DDPO models, we used an effective batch size of 8. When generating images with height and width 64, 128, and 512, we set the batch size to 8 without using gradient accumulation \citep{kozodoi2021gradient}. Curiously, for images of height and width 256, using a batch size of 8 caused an error: \textbf{RuntimeError: CUDA error: CUBLAS\_STATUS\_ALLOC\_FAILED when calling cublasCreate(handle)}. Thus, we opted to use a batch size of 4 with 2 gradient accumulation steps, equivalent to an effective batch size of 8, which worked. In order to investigate this error, we tried training a DDPO model with a batch size of 8 on a slower CPU, which was allocated 64 GB of memory. On the CPU, the error disappeared. We conclude that the reason for this error is dependent on how exactly variables are allocated across GPUs, but a more thorough investigation is beyond the scope of this paper.

\begin{table}[h!]
\centering
\begin{tabular}{||c c||} 
 \hline
 Hyperparameter & Value  \\ [0.5ex] 
 \hline\hline
 Epochs & 100 \\ 
 Batch Size & 32 \\
 Optimizer & Adam \\
 Learning Rate & 0.001 \\
  Adam \(\beta_1\) & 0.9 \\
 Adam \(\beta_2\) & 0.99 \\
  Adam Weight decay & 0.0 \\
 Adam \(\epsilon\) & 1e-8 \\
 Noise Dim & 100 \\
 Wasserstein \(\lambda\) & 10 \\
 Leaky ReLU negative slope & 0.2 \\
 Convolutional Kernel & 4 \\
 Convolutional Stride & 2 \\
 Transpose Convolutional Kernel & 4 \\
 Transpose Convolutional Stride & 2\\
 \hline
\end{tabular}
\caption{CAN Hyperparameters}
\label{tab:can-hyper}
\end{table}

\begin{table}[h]
    \centering
    \begin{tabular}{|| c | c | c||}
    \hline
    Model & Hours & kgCO$_2$ \\
    \hline
       CAN (Image Dim 512) & 121.39 & 13.11 \\
       Diffusion- CLIP Based (Image Dim 512)  & 22.75 & 2.46 \\
       Diffusion- K-Means Text Based (Image Dim 512)  & 21.50 & 2.32 \\
       Diffusion- K-Means Image Based (Image Dim 512) & 21.50 & 2.32 \\
       Diffusion- DCGAN Based (Image Dim 512) & 21.44 & 2.32 \\
         \hline
        CAN (Image Dim 256) & 72.94 & 7.88 \\
       Diffusion- CLIP Based (Image Dim 256)  & 11.2 & 1.21 \\
       Diffusion- K-Means Text Based (Image Dim 256)  & 17.25 & 1.86 \\
       Diffusion- K-Means Image Based (Image Dim 256) & 12.67 & 1.37 \\
       Diffusion- DCGAN Based (Image Dim 256) & 8.53  & 0.92 \\
         \hline
        CAN (Image Dim 128) & 66.14 & 7.15 \\
       Diffusion- CLIP Based (Image Dim 128)  & 7.05 & 0.76 \\
       Diffusion- K-Means Text Based (Image Dim 128)  & 6.40 & 0.69 \\
       Diffusion- K-Means Image Based (Image Dim 128) & 6.65 & 0.72 \\
       Diffusion- DCGAN Based (Image Dim 128) & 6.24  & 0.67 \\
         \hline
         CAN (Image Dim 64) & 83.33  & 9 \\
       Diffusion- CLIP Based (Image Dim 64)  & 7.30 & 0.79 \\
       Diffusion- K-Means Text Based (Image Dim 64)  & 6.85 & 0.74 \\
       Diffusion- K-Means Image Based (Image Dim 64) & 6.55 & 0.71 \\
       Diffusion- DCGAN Based (Image Dim 64) & 6.14  & 0.66 \\
         \hline
    \end{tabular}
    \caption{Training}
    \label{tab:training}
\end{table}

\subsection{Architecture}
For diffusion model training, the text encoder, autoencoder and unet were all loaded from \url{https://huggingface.co/stabilityai/stable-diffusion-2-base}. These model components were all frozen, but we added trainable LoRA weights to the cross-attention layers of the Unet. Parameter counts are shown in table \ref{tab:parameters}. The diffusion model components used the same amount of parameters regardless of image size, but the generator and discriminator had more parameters as image size increased.

\begin{table}[]
    \centering
    \begin{tabular}{||c|c|c|c||}
    \hline
    Model Component & Total Parameters & Trainable Parameters & Percent Trainable \\
    \hline
    Text Encoder & 34,0387,840 & 0 & 0\%  \\
    Autoencoder & 83,653,863 & 0 & 0\% \\
    UNet & 866,740,676 & 829,952 & 0.1\% \\
    \hline
    Generator (Image Dim 64) & 47,336,960 & 47,336,960 & 100\% \\
    Discriminator (Image Dim 64) & 13,691,612 & 13,691,612 & 100\% \\
    Generator (Image Dim 128) & 47,855,360 & 47,855,360 & 100\%\\
    Discriminator (Image Dim 128) & 14,347,228 & 14,347,228 & 100\% \\
    Generator (Image Dim 256) & 47,983,488 & 47,983,488 & 100\% \\
    Discriminator (Image Dim 256) & 15,920,604 & 15,920,604 & 100\% \\
    Generator (Image Dim 512) & 48,014,784 & 48,014,784 & 100\% \\
    Discriminator (Image Dim 512) & 20,115,932 & 20,115,932 & 100\%\\
    \hline
    \end{tabular}
    \caption{Parameter Counts}
    \label{tab:parameters}
\end{table}

\begin{figure}[h]
    \centering
    \includegraphics[scale=0.75]{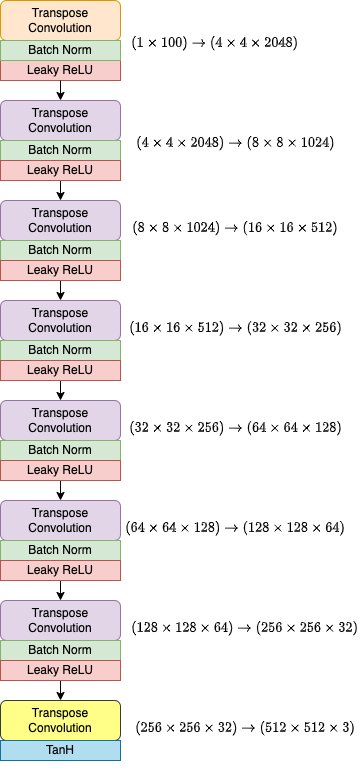}
    \caption{Generator Architecture (Image Dim 512)}
    \label{fig:gen-arc512}
\end{figure}

\begin{figure}[h]
    \centering
    \includegraphics[scale=0.75]{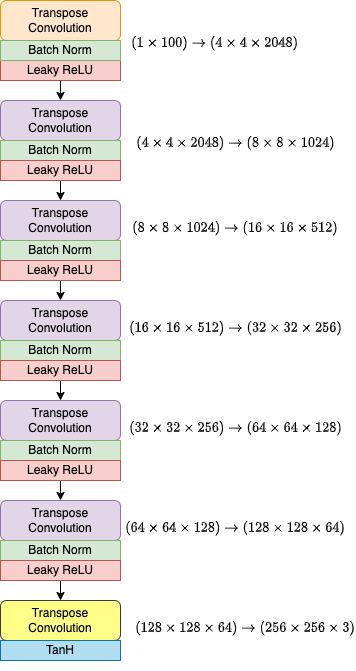}
    \caption{Generator Architecture (Image Dim 256)}
    \label{fig:gen-arc256}
\end{figure}

\begin{figure}[h]
    \centering
    \includegraphics[scale=0.75]{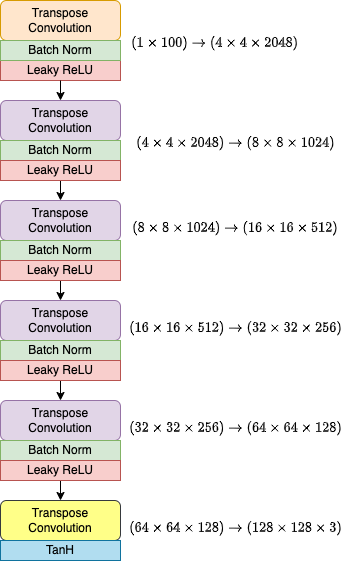}
    \caption{Generator Architecture (Image Dim 128)}
    \label{fig:gen-arc128}
\end{figure}

\begin{figure}[h]
    \centering
    \includegraphics[scale=0.75]{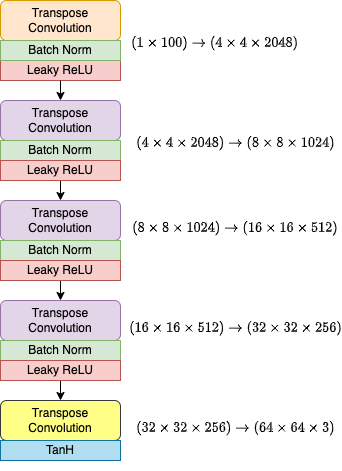}
    \caption{Generator Architecture (Image Dim 64)}
    \label{fig:gen-arc64}
\end{figure}

We used the convolutional neural network \citep{dumoulin2018guide} architecture described in \citet{ElgammalLEM17} for the CAN but had to use more/less layers to produce higher/lower dimension images. The generator takes a \(1 \times 100\) gaussian noise vector \(\in \mathbb{R}^{100} \sim \mathcal{N}(0, I)\) and maps it to a \(4 \times 4 \times 2048\) latent space, via a convolutional transpose layer with kernel size  = 4 and stride =1, followed by 3, 4, 5 or 6 transpose convolutional layers corresponding to image dimensions 64, 128, 256 and 512, each upscaling the height and width dimensions by two, and halving the channel dimension (for example one of these transpose convolutional layers would map \(\mathbb{R}^{4 \times 4 \times 2048} \rightarrow \mathbb{R}^{8 \times 8 \times 1024}\)) followed by batch normalization \citep{ioffe2015batch} and Leaky ReLU \citep{maas2013rectifier}, and then one final convolutional transpose layer with output channels = 3 and tanh \citep{dubey2022activation} activation function. Diagrams of the generators with image dim 512, 256, 128 and 64 are shown in the figures 
 \ref{fig:gen-arc512}, \ref{fig:gen-arc256}, \ref{fig:gen-arc128} and \ref{fig:gen-arc64} respectively.

\begin{figure}[h]
    \centering
    \includegraphics[scale=0.6]{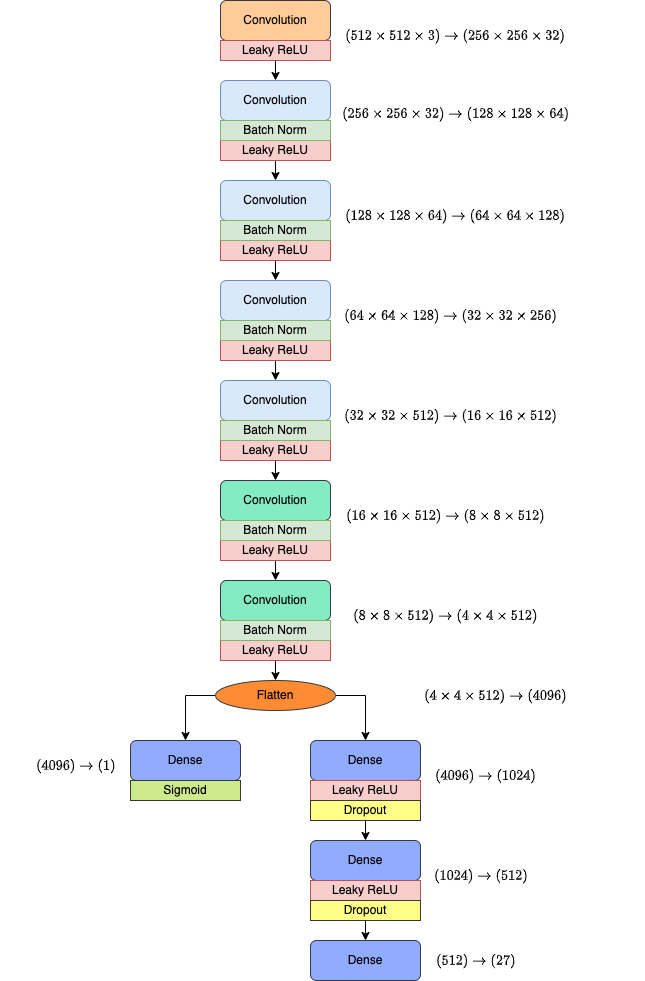}
    \caption{Discriminator Architecture (Image Dim 512)}
    \label{fig:disc-arc512}
\end{figure}

\begin{figure}[h]
    \centering
    \includegraphics[scale=0.6]{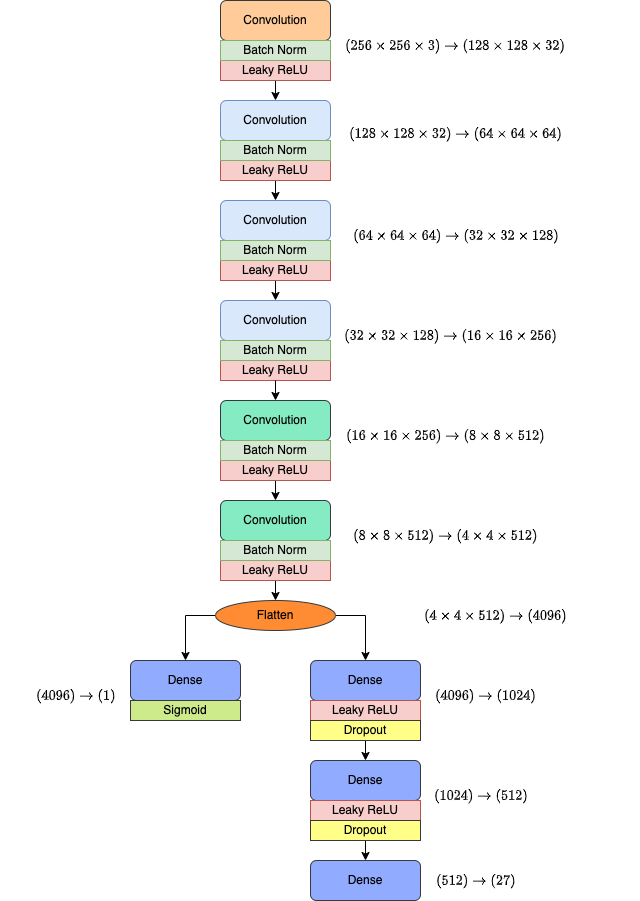}
    \caption{Discriminator Architecture (Image Dim 256)}
    \label{fig:disc-arc256}
\end{figure}

\begin{figure}[h]
    \centering
    \includegraphics[scale=0.65]{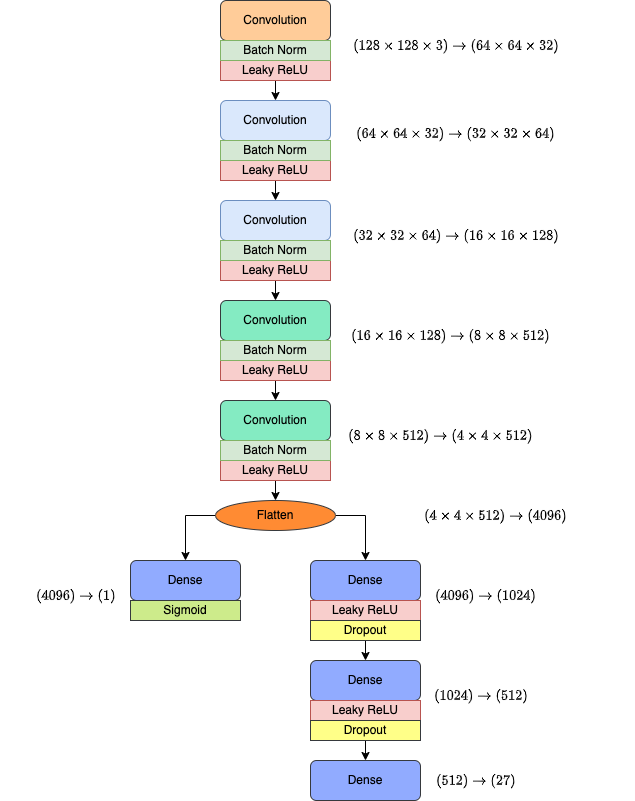}
    \caption{Discriminator Architecture (Image Dim 128)}
    \label{fig:disc-arc128}
\end{figure}

\begin{figure}[h]
    \centering
    \includegraphics[scale=0.65]{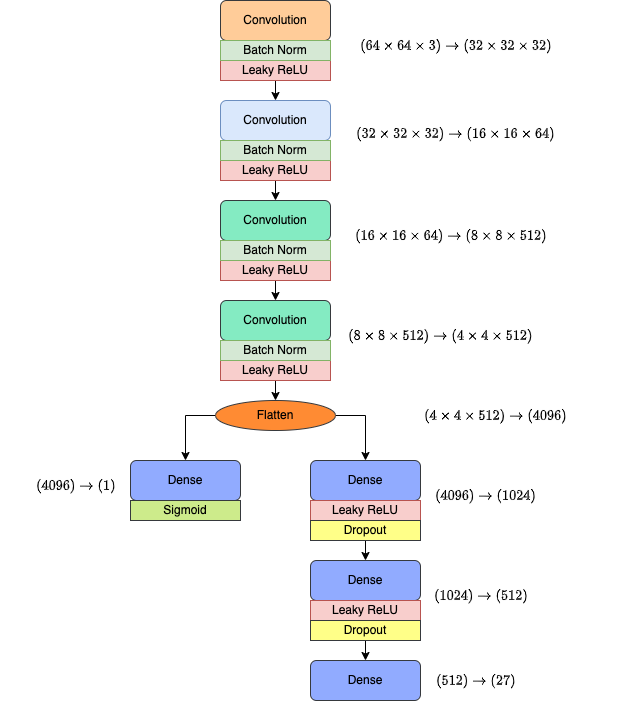}
    \caption{Discriminator Architecture (Image Dim 64)}
    \label{fig:disc-arc64}
\end{figure}

For the discriminator, we first applied a convolution layer to downscale the input image height width dimensions by 2 and mapped the 3 input channel dimensions to 32 (\(\mathbb{R}^{512 \times 512 \times 3} \rightarrow \mathbb{R}^{256 \times 256 \times 32}\)) with Leaky ReLU activation. Then we had 2, 3, 4 or 5 convolutional layers corresponding to image dimensions 64, 128, 256 and 512, each downscaling the height and width dimensions by 2 and doubling the channel dimension (for example, one of these convolutional layers would map \(\mathbb{R}^{256 \times 256 \times 32} \rightarrow \mathbb{R}^{128 \times 128 \times 64}\)) with batch normalization and Leaky ReLU activation. Then we had two more convolutional layers, each downscaling the height and width dimensions but keeping the channel dimensions constant (using the prior layer's channel dimensions), with batch normalization and Leaky ReLU activation. The output of the convolutional layers was then flattened. The discriminator had two heads- one for style classification (determining which style a real image belongs to) and one for binary classification (determining whether an image was real or fake). The binary classification head consisted of one linear layer with one output neuron. The style classification layer consisted of 2 linear layers with LeakyReLU activation and Dropout, with output 1024 output neurons and 512 output neurons, respectively, followed by a linear layer with 27 output neurons for the 27 artistic style classes. Diagrams of discriminators with image dim 512, 256, 128, and 64 are shown in the figures \ref{fig:disc-arc512}, \ref{fig:disc-arc256}, \ref{fig:disc-arc128}, and \ref{fig:disc-arc64}, respectively.

\end{document}